\documentclass{aamas2015}

\usepackage{subfigure}
\usepackage[justification=centering]{caption}
\usepackage{multirow}

\usepackage{floatrow}
\newfloatcommand{capbtabbox}{table}[][\FBwidth]

\newcommand{\expE}{\mathbb E}
\DeclareMathOperator*{\argmax}{arg\,max}

\makeatletter
\let\@copyrightspace\relax
\makeatother

\begin{document}
\title{Off-Policy Reward Shaping with Ensembles}

\numberofauthors{1}

\author{
\alignauthor
%Paper  34
Anna Harutyunyan, Tim Brys, Peter Vrancx and Ann Now\'{e}\\
      %\affaddr{Artificial Intelligence Lab}\\
\vspace{1em}
      \affaddr{Vrije Universiteit Brussel}\\
      \email{\{aharutyu,timbrys,pvrancx,anowe\}@vub.ac.be}
% \alignauthor Tim Brys\\
%        %\affaddr{Artificial Intelligence Lab}\\
%       \affaddr{Vrije Universiteit Brussel}\\
%       \email{tbrys@vub.ac.be}
% \alignauthor Peter Vrancx\\
%        %\affaddr{Artificial Intelligence Lab}\\
%       \affaddr{Vrije Universiteit Brussel}\\
%       \email{pvrancx@vub.ac.be}
% \and \alignauthor Ann Now\'{e}\\
%        %\affaddr{Artificial Intelligence Lab}\\
%       \affaddr{Vrije Universiteit Brussel}\\
%       \email{anowe@vub.ac.be}
}

\maketitle

\begin{abstract}
Potential-based reward shaping (PBRS) is an effective and popular
technique to speed up reinforcement learning by leveraging domain
knowledge. While PBRS is proven to always preserve optimal policies, its
effect on learning speed is determined by the {\em quality} of its
potential function, which, in turn, depends on both the underlying
heuristic and the scale. Knowing which heuristic
will prove effective requires testing the options beforehand, and
determining the appropriate scale requires tuning, both of which
introduce additional sample complexity.

We formulate a PBRS framework that improves learning speed, but
does not incur extra sample
complexity. For this, we propose to {\em simultaneously} learn an ensemble of policies, shaped w.r.t. many
heuristics and on a range of scales. The target policy is
then obtained by voting. The ensemble needs to be able to efficiently
and reliably learn off-policy: requirements fulfilled by the recent
Horde architecture, which we take as our basis. We demonstrate
empirically that (1) our ensemble policy outperforms both the base policy,
and its single-heuristic components, and (2) an ensemble over a general range of scales
performs at least as well as one with optimally tuned
components.

% While one can typically think of many simple solution heuristics for a given
% problem, the challenge is knowing a priori which will be most
% effective in practice. Further, given a heuristic, choosing its
% scaling factor optimally requires tuning beforehand, which is problematic in practice, % a luxury
% % unavailable,
% when the environment samples are costly. .

% It is much more challenging to design a single universally effective
% heuristic, than many locally effective simpler ones. The challenge is
% then choosing amongst them, a typically infeasible task a priori.
% In this paper we suggest to leverage the strengths of {\em all} of them
% at once by maintaining policies w.r.t. each of them in an ensemble.
% On the other hand, determining the optimal scaling factor for a given
% heuristic, requires tuning. 

\end{abstract}

% \category{I.2.6}{Artificial
%   Intelligence}{Learning}% \category{I.2.11}{Artificial
%   % Intelligence}{Distributed Artificial Intelligence}[Intelligent agents]

% %A category including the fourth, optional field follows...
% %\category{D.2.8}{Software Engineering}{Metrics}[complexity measures, performance measures]

% %General terms should be selected from the following 16 terms: Algorithms, Management, Measurement, Documentation, Performance, Design, Economics, Reliability, Experimentation, Security, Human Factors, Standardization, Languages, Theory, Legal Aspects, Verification.

% \terms{Algorithms, Experimentation}

% %Keywords are your own choice of terms you would like the paper to be indexed by.

% \keywords{reinforcement learning, potential-based reward shaping, off-policy learning, horde}

\section{Introduction}
\label{sec:introduction}

The powerful ability of reinforcement learning (RL)~\cite{sutton-barto98} to find optimal
policies {\em tabula rasa}, is also the source of its main
weakness: infeasibly long running times. As the
problems RL tackles get larger, it becomes increasingly important to
leverage all possible knowledge about the domain at hand. One paradigm to
inject such knowledge into the reinforcement learning problem is {\em
  potential-based reward shaping (PBRS)}~\cite{ng99}. Aside from
repeatedly demonstrated 
efficacy in increasing learning
speed~\cite{asmuth2008potential,devlin11,brys2014combining,snelshimonshaping},
the principal strength of PBRS lies in its ability to preserve optimal policies. Moreover, it
is the only\footnote{Given no knowledge of the environment dynamics.}
reward shaping scheme that is guaranteed to do so~\cite{ng99}. At the
heart of PBRS methods lies the {\em potential function}. Intuitively,
it expresses the ``desirability'' of a state, defining the {\em
  shaping reward} on a transition to be the {\em
  difference} in potentials of the transitioning states. States may be
desirable by many criteria. The pursuit of designing a potential function that
accurately encapsulates the ``true'' desirability is meaningless, as
it would solve the task at hand~\cite{ng99}, and remove the need for
learning altogether. However, one can usually suggest many simple
heuristic criteria that improve performance in different
situations. Choosing the most effective heuristic amongst them
without a test comparison, is typically infeasible, and carrying out
such a comparison implies
added sample complexity, that may be unaffordable. Moreover, heuristics may contribute
complementary knowledge that cannot be leveraged in
isolation~\cite{brys2014combining}.

The {\em choice} of a heuristic is merely one of the two deciding factors for the
performance of a potential function. The other (and one that is even less
intuitive) is {\em scaling}. An
effective heuristic with a sub-optimal scaling factor may make no difference
at all, if the factor is too small, or dominate the base reward and
distract the learner,\footnote{The agent will eventually
  still uncover the optimal policy, but instead of helping him get
  there faster, reward shaping would slow the learning down.} if the
factor is too large. %  This is an instance of the {\em magnitude scaling problem},
% % ubiquitous
% commonly arising in value-based RL methods, particularly those combining
% heterogeneous values% ~\cite{???}
% . 
Typically, one is required to tune
the scaling factor beforehand, which requires extra environment
samples, and is infeasible in realistic problems.

We wish to devise a PBRS framework that is capable of improving
learning speed, without introducing extra sample % or computational
complexity. % Inspired by the principle behind ensemble methods% to obtain accurate
% % classifiers by combining many na\"ive inaccurate ones
% , 
To this end, rather than learn a single policy shaped with the most
effective heuristic on its optimal scale, we propose to maintain an {\em
  ensemble} of policies that all learn from the same experience, but are shaped w.r.t. different
heuristics and different scaling factors. The deployment of our ensemble thus does not require any
additional environment samples, and frees the designer up to benefit
from PBRS, equipped only with a set of intuitive heuristic rules, with
no necessary knowledge of their performance and value magnitudes. 

Because (for the purpose of not requiring extra environment samples),
all member-policies learn to maximize different reward functions from the
same experience, the learning needs to be reliable {\em
  off-policy}. Because the introduced computational
complexity (for each of the additional member-policies) amounts to
that of the off-policy learner, we wish
for the learning to be as efficient as possible. The recently introduced {\em
  Horde} architecture~\cite{sutton09} is well-suited to be the basis
of our ensemble, due to its general off-policy convergence
guarantees and computational efficiency. In contrast to the previous uses of
Horde~\cite{pilarski2013}, we exploit its power to learn a {\em
  single} task, but from multiple viewpoints.

The convergence guarantees of Horde require a {\em latent} learning
scenario~\cite{maei2010}, i.e. one of (off-policy) learning under a
fixed (or slowly changing) behavior policy. This scenario is particularly relevant to
real-world applications, where failure is highly penalized and the
usual trial-and-error tactic is implausible, e.g. robotic setups. One
could imagine the agent following a safe exploratory policy, while
learning the target control policy, and only executing the target
policy after it is learnt. That is the scenario we focus on in this
paper. Note that the conventional interpretation of PBRS to steer
exploration~\cite{grzes2010diss}, does not apply here, as the behavior
is unaffected by the target policy, and is kept fixed.  This work (and
its precursor~\cite{harutyunyan2014ecai}) provides, to our knowledge,
the first validation of PBRS effective in such a latent setting.

Our contribution is two-fold: (1) we formulate and empirically validate a PBRS
framework as a policy ensemble, that is capable of increasing learning
speed without
adding extra sample complexity, and that does so with general convergence
guarantees. Specifically, we demonstrate how such an ensemble can be
used to lift the problems of both the {\em choice} of the potential
function and its {\em scaling}, thus removing the need of
behind-the-scenes tuning necessary before deployment; and (2) we validate PBRS to be effective in
a {\em latent} off-policy setting, in which it cannot steer the exploration strategy.

% \paragraph{Outline}  
In the following section we give an overview of the
preliminaries. Section~\ref{sec:parallel-shaping} motivates our
approach further, while Section~\ref{sec:our-architecture} describes the
proposed architecture and the voting techniques used to obtain the
target ensemble policy. Section~\ref{sec:experimental-results}
presents empirical results in two classical benchmarks, and Section~\ref{sec:concl-future-work} concludes. % and discusses future work directions.

\section{Background}
\label{sec:background}

We assume the usual RL
framework~\cite{sutton-barto98}, in which the {\em agent} interacts
with its (typically) Markovian {\em environment} at discrete time steps
$t=1,2,\ldots$. Formally, a {\em Markov Decision
  Process (MDP)}~\cite{puterman94} is a tuple ${\cal M}=\langle
{\cal S}, {\cal A},\gamma, {\cal T}, R\rangle$, where: $\cal S$ is a
set of {\em states}, $\cal A$ is a set of {\em actions}, $\gamma\in
[0,1]$ is the {\em discounting factor}, ${\cal
  T}=\{P_{sa}(\cdot)|s\in{\cal S}, a\in{\cal A}\}$ are the next state
{\em transition probabilities} with $P_{sa}(s')$ specifying the
probability of state $s'$ occuring upon taking action $a$ from state
$s$, $R: {\cal
  S} \times {\cal A} \times {\cal S} \rightarrow {\mathbb R}$ is the
{\em reward function} with $R(s,a,s')$ giving
the expected value of the reward that will be received when $a$ is
taken in state $s$, and $r_{t+1}$ denoting the component of $R$ at time
$t$.

A (stochastic) Markovian {\em policy} $\pi : {\cal
  S} \times {\cal A}\rightarrow [0,1]$ is a probability
distribution over actions at each state, s.t.
$\pi(s,a)$ gives the probability of action $a$ being taken from state
$s$ under policy $\pi$. In the deterministic case, we will take $\pi(s)=a$ to mean
$\pi(s,a)=1$. 

Value-based methods encode policies
through {\em value functions}, which denote expected cumulative reward
obtained while
following the policy. We focus on {\em state-action} value functions. In a discounted setting:

\begin{equation}
  \label{eq:20}
  Q^{\pi}(s,a) = \expE_{{\cal
      T},\pi}\Big[\sum\limits_{t=0}^{\infty}\gamma^t
  r_{t+1}|s_0=s,a_0=a\Big]
\end{equation}

An action $a^*$ is {\em greedy} in a state $s$, if it is the action of maximum
value in $s$. A (deterministic) policy is greedy, if it picks the
greedy action in each state:

% (deterministic) {\em greedy} policy is obtained by picking the action of maximum value at each state:

\begin{equation}
  \label{eq:1}
  \pi(s)=\argmax_{a} Q^{\pi}(s,a), \forall s\in {\cal S}
\end{equation}

A policy $\pi^*$ is {\em optimal} if its value is largest:

\begin{equation*}
  \label{eq:28}
  Q^*(s,a)=\sup_{\pi} Q^{\pi}(s,a), \forall s\in{\cal S}, \forall a\in{\cal A}
\end{equation*}

%On-policy learning evaluates the policy being followed, whereas 

The learning is {\em on-policy} if the {\em behavior} policy $\pi_b$ that the agent
is following is the same as the {\em target} policy $\pi$ that the
agent is evaluating. Otherwise, it is {\em off-policy}.  Given $\pi_b$, the values of the optimal greedy policy can be learned incrementally
through the following {\em Q-learning}~\cite{watkins1992} update:

\begin{equation}
  \label{eq:19}
  Q_{t+1}(s_t,a_t) = Q_t(s_t,a_t) + \alpha_t\delta_t
\end{equation}

\begin{equation}
  \label{eq:8}
  \delta_t=r_t+\gamma\max\limits_{a^*\in{\cal A}} Q_t(s_{t+1},a^*)-Q_t(s_t,a_t)
\end{equation}

where $Q_t$ is an estimate of $Q^{\pi}$ at time $t$, $\alpha_t\in
(0,1)$ is the {\em learning rate} at time $t$, $a_t$ is chosen
according to $\pi_b$, $\delta_t$ is the {\em temporal-difference
  (TD) error} of the transition. $s_{t+1}$ is drawn according to ${\cal T}$,
given $s_t$ and $a_t$, and $a^*$ is the greedy action
w.r.t. $Q_t$ in $s_{t+1}$ % This update is known as
% {\em Q-learning}~\cite{watkins1992}, arguably the most popular off-policy RL
% algorithm. 
Given tabular representation, this process is shown to converge to the
correct value estimates (the {\em TD-fixpoint}) in the limit under standard approximation conditions~\cite{jaakkola1994convergence}.

When the state or action spaces are too large, or continuous, tabular
representations do not suffice and one needs to use function
approximation (FA). The state (or state-action) space is then
represented through a set of features $\phi$, and the algorithms learn
the value of a parameter vector $\theta$. In the (common) linear case:

\begin{equation}
  \label{eq:5}
  Q_t(s,a)=\theta_t^T\phi_{s,a},\forall s\in {\cal S},\forall a\in
  {\cal A}
\end{equation}

and Eq.~\eqref{eq:19} becomes:

\begin{equation}
  \label{eq:11}
  \theta_{t+1}  = \theta_t + \alpha_t\delta_t\phi_t,
\end{equation}

where we slightly abuse notation by letting $\phi_t$ denote the state-action
features $\phi_{s_t,a_t}$, and $\delta_t$ is still computed according
to Eq.~\eqref{eq:8}.

In the next two subsections we present the core ingredients
to our approach.

\subsection{Horde} 

% Unfortunately, 
FA is known to cause off-policy bootstrapping methods (such as
Q-learning) to diverge even on simple
problems~\cite{baird95,Tsitsiklis97ananalysis}. The family of {\em
  gradient temporal difference (GTD)} methods provides a solution for
this issue, and guarantees off-policy convergence under FA, given a fixed (or slowly
changing behavior)~\cite{sutton09}. Previously, similar guarantees
were provided only by second-order {\em batch} methods (e.g. LSTD~\cite{Bradtke96linearleast-squares}),
unsuitable for online learning. GTD methods are the first to maintain
these guarantees, while maintaining the (time and space) complexity linear in the size of the state
space. Note that linearity is a lower bound on what is achievable, because it is required to simply store and access the
learning vectors. As a consequence, GTD methods scale well to the
number of value functions (policies)
learnt~\cite{modayil2012acquiring}, and due to the inherent off-policy
setting, can do so from a single stream of environment interactions
(or {\em experience}). Sutton et al.~\cite{sutton11} formalize this idea in
a framework of parallel off-policy learners, called {\em Horde}. They demonstrate Horde to
be able to learn thousands of predictive and goal-oriented value
functions in real-time from a single unsupervised stream of sensorimotor experience. There have been further successful
applications of Horde in realistic robotic
setups~\cite{pilarski2013}. 

On the technical level,\footnote{Please
refer to Maei's dissertation for the full details~\cite{maei-diss}.} GTD methods are based on the idea of
performing gradient descent on a reformulated objective function,
which ensures convergence to the {\em projected} TD-fixpoint, by
introducing a gradient bias into the
TD-update~\cite{sutton09}. Mechanistically, it requires maintaining
and learning a second set of weights $w$, along with $\theta$, and performing the following
updates:% \footnote{This is the simplest form of the update rules for
  % gradient temporal-difference algorithms, namely that of
  % TDC~\cite{sutton09}. GQ($\lambda$)~\cite{maei2010gq} augment this update with eligibility traces.}

\begin{eqnarray}
  \label{eq:12}
  \theta_{t+1} & = &  \theta_t+\alpha_t\delta_t\phi_t-\alpha\gamma\phi'_{t}(\phi_t^T
  w_t)\\
  w_{t+1}    & = &  w_t+\beta_t(\delta_t-\phi_t^T w_t)\phi_t
\end{eqnarray}

where $\delta_t$ is still computed with Eq.~(\ref{eq:8}), and
$\phi'_t$ is the feature vector of the next state and action. This is
a simpler form of the GTD-update, namely that of
  TDC~\cite{sutton09}. GQ($\lambda$)~\cite{maei2010gq} augments this
  update with eligibility traces.

Convergence is one of the two theoretical hurdles with off-policy
learning under FA. The other has to do with the {\em quality} of
solutions under off-policy sampling, which may, in general, fall far from
optimum, even when the approximator can represent the true value
function well. In, to our knowledge, the only work that addresses this
issue, Kolter~\cite{kolter2011fixed} gives a way of constraining the solution
space to achieve stronger qualitative guarantees, but his
algorithm has quadratic complexity and thus is not scalable. Since
scalability is crucial in our framework, Horde remains the only
plausible convergent architecture available.

  \subsection{Reward Shaping} 
\label{sec:reward-shaping}

{\em Reward shaping} augments the true
  reward signal $R$ with an additional {\em shaping} reward $F$, provided by the
  designer. The shaping reward is intended to guide the agent, when
  the environmental rewards are sparse or uninformative, in order to
  speed up learning. In its most general form: % It was originally thought of as a way of scaling up
  % RL methods to handle difficult problems~\cite{Dorigo97robotshaping}, as RL generally suffers
  % from infeasibly long learning times.

  \begin{equation}
    \label{eq:16}
    R' = R+F
  \end{equation}

  Because tasks are identified by their reward function, modifying the
  reward function needs to be done with care, in order to not alter the
  task, or else reward shaping can slow down or even prevent
  finding the optimal policy~\cite{Randløv98}. Ng et al.~\cite{ng99}
  show that grounding the shaping rewards in {\em state potentials} is
  both necessary and sufficient for ensuring preservation of the
  (optimal) policies of the original MDP. {\em Potential-based reward
    shaping (PBRS)} maintains a potential function $\Phi : S \rightarrow
  {\mathbb R}$, and defines the auxiliary reward function $F$ as: 
\begin{equation}\label{eq:6}
  F(s,a,s')=\gamma\Phi(s')-\Phi(s)
\end{equation}

  where $\gamma$ is the discounting factor of the MDP. We refer to the rewards, value functions
and policies, augmented with shaping rewards as {\em shaped}. Shaped
policies converge to the same (optimal) policies as the base learner, but differ during the learning process.

\section{A Horde of Shapings}
\label{sec:parallel-shaping}

% In this section we further discuss PBRS as a source of components for
% effective policy ensembles in RL, and Horde as a well-suited framework for maintaining such an ensemble.

% believe Horde to be well-suited
%  for expressing ensembles in RL, when one's aim is to reduce learning
%  speed without incurring extra sample costs. We then
%  further motivate our choice of PBRS as a
%  source of components for such an ensemble.

The key insight in ensemble learning is that the strength of an
ensemble lies in the {\em diversity} its components
contribute~\cite{Krogh95neuralnetwork}. In the RL context, this diversity can
be expressed through several aspects, related to dimensions of the learning process: (1)
diversity of {\em experience}, (2) diversity of {\em algorithms} and (3)
diversity of {\em reward signals}. 
Diversity of experience naturally implies high sample complexity, and
assumes either a multi-agent setup, or learning in stages. Diversity
of algorithms (given the same experience) is computationally costly, as it requires separate
representations, and one needs to be particular about the choice of
algorithms due to convergence considerations.\footnote{See the
  discussion on convergence in Section 6.1.2 of van Hasselt's
  dissertation~\cite{hasselt2011}.} In
the context of our aim of increasing learning speed, without
introducing complexity elsewhere, we focus on the latter
aspect of diversity: diversity of reward signals. 

PBRS is an elegant and theoretically attractive approach to
introducing diversity into the reward function, by drawing from the
available domain knowledge. Such knowledge can often be described as a set of simple heuristics. Combining the
corresponding potentials beforehand na\"ively (e.g. with linear scalarization) may result in
information loss, when the heuristics counterweigh each other, and
introduce further scaling issues, since the relative magnitudes of
the potential functions may differ. Maintaining the shapings
separately has recently been shown to be a more robust and effective
approach~\cite{brys2014combining}. Under the requirements of
convergence and efficiency, maintaining such an ensemble of policies
learning in parallel and shaped with different potentials, is only
possible via the Horde architecture, which is the approach we take in this
paper. Thus, the proposed ensemble is the first of its kind to possess
general convergence guarantees.%  Note that this argument is not limited to PBRS ensembles. An
% off-policy setup seems inevitable if one wants to design an effective policy ensemble: the required diversity
% is unlikely to be found in the behavior policy alone, and
% member-policies must contribute information different from the
% behavior. Thus, Horde is well-suited to express policy ensembles in general.

Horde's demonstrated ability to learn thousands of policies
in parallel in real time~\cite{sutton11,modayil2012acquiring} allows to consider large
ensembles, at little computational cost. While defining thousands of distinct
heuristics is rarely sensible, each heuristic may be learnt on
many different scaling factors. This not only frees one from having to
tune the scaling factor a priori (one of the issues we focus on in this paper), but potentially allows for
automatically dynamic scaling, corresponding to {\em state-dependent}
shaping magnitudes.

\subsubsection*{Shaping Off-Policy}
\label{sec:shaping-policy}

The effects of PBRS on the learning process are usually considered to lie in the
guidance of exploration during
learning~\cite{grzes2010diss,marthi2007,ng99}. Laud and
DeJong~\cite{laud03} formalize this by showing that the difficulty of
learning is most dependent on the {\em reward horizon}, a measure of the
number of decisions a learning agent must make before experiencing
accurate feedback, and that reward shaping artificially reduces this
horizon. In our latent setting we assume no control over the agent's
behavior. The performance benefits then can be
explained by faster {\em knowledge propagation} through the
TD updates, which we now observe decoupled from
guidance of exploration.

 Reward shaping in such off-policy settings is not well studied or
understood, and these effects are of independent interest.

% \subsection*{RL Ensembles via Horde}
% \label{sec:rl-ensembles-via}

% Most previous uses of policy ensembles in RL involved independent
% runs for each policy, with the combination happening post-factum~\cite{fausser2011}. This is limited in practical utility,
% since it requires a large computational and sample overhead,
% assumes a repeatable setup, and does not improve learning
% speed. Others, in general, lack convergence guarantees, by either using mixed on- and off-policy
% learners~\cite{wiering08}, or Q-learners under
% FA~\cite{brys2014combining}. 
% Generally, when considering policy ensembles in RL, an off-policy learning setup
% seems inevitable, if one wants to be effective: the required diversity
% is unlikely to be found in the behavior policy alone, and
% member-policies must contribute information different from the
% behavior. Horde is the first framework that allows to efficiently and
% reliably learn multiple policies off-policy, making it a well-suited
% basis for ensemble learning in RL.

\section{Architecture}
\label{sec:our-architecture}

\begin{figure}[h]
  \centering
  \includegraphics[scale=0.5]{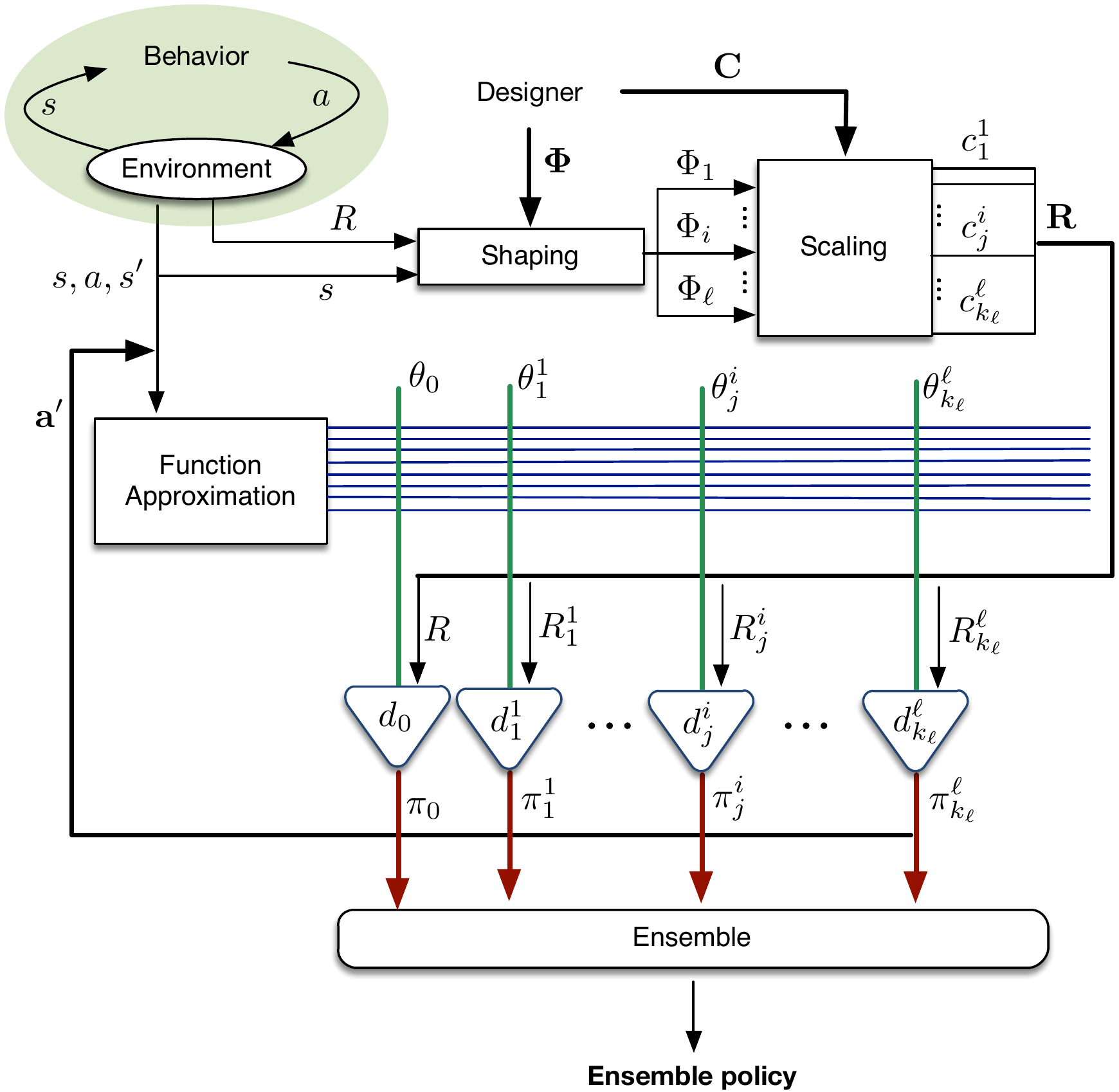}
  \caption{\small An overview of the Horde
    architecture used to learn an ensemble
    of shapings (including the base learner). Vectors are indicated with bold lines. $R^i_j$ is the reward obtained when applying $\Phi_i$
    to $R$ and scaling with $c^i_j$. The blue output of
    the linear function approximation block are the features of the
    transition (two state-action pairs), with their intersections with 
    $\theta^i_j$ representing weights. ${\bf a'}$ is
    a vector of greedy actions at $s'$ w.r.t. to each policy
    $\pi^i_j$. Note that in this latent settings, all interactions with the environment happen only in the upper left corner.}
  \label{fig:architecture}
\end{figure}

We are now ready to describe the architecture of our ensemble (Fig.~\ref{fig:architecture}).% \footnote{Figure idea from Sutton's presentation~\cite{sutton-horde-slides}} 
We maintain our Horde of shapings as a set $\cal D$ of Greedy-GQ($\lambda$)-learners~\cite{maei2010gq}. Given a set of potential functions
${\bf \Phi}=\{\Phi_1,\ldots \Phi_{\ell}\}$ a range of scaling
factors ${\bf c}^i=\langle c^i_1,\ldots c^i_{k_i}\rangle$ for each
$\Phi_i$, and the base reward function $R$, the ensemble reward function is a vector:

\begin{equation}
  \label{eq:15}
  {\bf R}=R+\langle F_{c^1_1}^{\Phi_1},F_{c^1_2}^{\Phi_1},\ldots,F_{c^{\ell}_{k_{\ell}}}^{\Phi_{\ell}}\rangle
\end{equation}

where $F^{\Phi_i}_{c^i_{j}}$ is the potential-based shaping reward given by
  Eq.~\eqref{eq:6} w.r.t. the potential function $\Phi_i$ and scaled
  with the factor $c^i_j$. For notational clarity, we will take
  $F^i_j$ to mean $F^{\Phi_i}_{c^i_{j}}$ (i.e. the shaping w.r.t. to
  the $i$-th potential function on the $j$-th scaling factor), and
  $R^i_j=R+F^i_j$. We allow the ensemble the option to include the
  base learner.

%   $i=1,\ldots,|{\bf\Phi}|-1$ are
% the potential-based rewards given by (\ref{eq:6}) on potentials $\Phi_1,\Phi_2,\ldots$
% provided by the designer, and $j=1,\ldots,|C|-1$ are the scaling
% factors from the range $C$. 

We adopt the terminology of Sutton et
al.~\cite{sutton11}, and refer to individual agents within Horde as
{\em demons}. Each demon $d^i_j$ learns a greedy policy $\pi^i_j$ w.r.t. its
reward $R^i_j$. Recall that our latent setting implies that the learning is guided by a fixed behavior policy
$\pi_b$, with $\pi^i_j$ all learning in parallel from the experience
generated by $\pi_b$. Because each policy $\pi^i_j$ is available
separately at each step, an {\em ensemble}
policy can be devised by collecting votes on action preferences from
all demons $d^i_j$. The ensemble is also latent, and not executed
until the learning has ended. Note that because PBRS preserves {\em all} of the optimal
policies from the original problem~\cite{ng99}, the ensemble policy does too.

In this paper we have considered two voting schemes: {\em majority} voting and {\em rank}
voting, which are elaborated below. The architecture is certainly not limited to these choices.
% (or Borda count~\cite{???})

% Recall that the learning is latent, and
% guided by a fixed behavior policy $\pi_b$, with all $\pi^i_j$ learning
% from the experience generated by $\pi_b$ at the same time. The
% ensemble policy can be constructed either latently during learning, or executed after the learning has stopped. 

% A note on optimality. PBRS preserves {\em all} of the optimal
% policies from the original problem. This is because the optimal value
% function of $M'$ is simply offset from that of $M$ by the potential
% function a transformation invariant to policies~\cite{ng99}. 
% In practice, it is possible that a
% potential-guided policy finds {\em some} optimal policy first, before
% all the others. Which policy that is would be conditional on the
% shaping. It is thus possible that before the demons have converged to
% the true value function, individual demon's policies may be optimal,
% while the ensemble policy -- not yet. We do not expect this to be an
% issue in practice.

%PROVE THAT ENSEMBLE POLICY IS OPTIMAL??

% Note that, because the individual policies in the ensemble converge to the
% optimal policy by the guarantees of PBRS, the ensemble policy does as
% well. PROVE THIS? NOTE THAT SAME VALUE FUNCTION OFFSET BY DIFF POTENTIALS

\subsection{Ensemble Policy}
\label{sec:ensemble-policy}

To the best of our knowledge, both voting methods were first used in the context of RL agents by Wiering
and Van Hasselt~\cite{wiering08}. In both methods, each demon $d$ casts a
vote $v_{d} : S\times A \rightarrow {\mathbb N}^0$, s.t. $v_d(s,a)$ is the
preference value of action $a$ in state $s$. % , and $\sum\limits_{a\in A}
% v_d(s,a)=1, \forall s\in S$.
The voting scheme then is defined for policies, rather than
value functions, which mitigates the
magnitude bias.\footnote{Note that even though the
  shaped {\em policies} are the same upon convergence -- the value
  functions are not.}  The ensemble policy acts greedily (with ties
broken randomly) w.r.t. the cumulative {\em preference} values $P$:

\begin{equation}
  \label{eq:1}
  P(s_t,a)=\sum\limits_{d\in\cal D} v_{d}(s_t,a),\forall a\in A
\end{equation}

The voting scheme determines the manner in which $v_d$ are assigned.

\begin{description}
  \item[Majority voting] Each demon $d$ casts a vote of 1 for its most preferred
action, and a vote of 0 for the others. I.e.:

\begin{equation}
  \label{eq:7}
v_{d}(s,a) = 
\begin{cases} 1 &\mbox{if } Q(s,a)=\max\limits_{a^*}Q(s,a^*) \\ 
                      0 & \mbox{otherwise.} 
\end{cases} 
\end{equation}
\item[Rank voting] Each demon greedily ranks its $n$ actions, from $n-1$ for  its most, to $0$ for its least preferred
actions. We slightly modify the formulation
from~\cite{wiering08}, by ranking Q-values, instead of policy
probabilities. I.e. $v_d(s,a) > v_d(s,a')$, if and only if $Q_d(s,a) >
Q_d(s,a')$.
\end{description}

\section{Experiments}
\label{sec:experimental-results}

We now present the empirical studies that
validate the efficacy of our ensemble architecture w.r.t. both the
choice of heuristic and the choice of scale. % \footnote{We remind the reader that
% while all policies eventually arrive at the same (optimal) solution,
% our focus is the time it takes them to get there.
% } 
We first consider the
scenario of choosing between heuristics, and evaluate an ensemble
consisting of shapings with appropriate scaling factors. The experiments show that the
ensemble policy performs at least as well as the best heuristic. We
then turn to the problem of scaling, and demonstrate that ensembles on
both narrow and broad ranges of scales perform at least as well as the
one w.r.t. the optimal scaling factors.

We carry out our experiments on two common benchmark problems. In both
problems, the behavior policy is a uniform distribution over all
actions at each time step. The evaluation is done by interrupting the
base learner every $z$ episodes and executing the queried greedy policy once. No learning is allowed
during evaluation.

We evaluated the ensembles w.r.t. both voting schemes from
Sec.~\ref{sec:ensemble-policy}, and found the (sum) 
performance to be not
significantly different ($p > 0.05$), with {\em rank} voting
performing slightly better. To keep the clarity of focus, % and respect
% the space considerations, 
below we only present the results for the
rank voting scheme, but emphasize that the performance is not
conditional on this choice.

\subsection{Mountain Car}
\label{sec:mountain-car}

We begin with the classical benchmark domain of mountain
car~\cite{sutton-barto98}. The task is to drive an underpowered car up a hill.
The (continuous) state of the system is composed of the current position (in $[-1.2,0.6]$)
and the current velocity (in $[-0.07,0.07]$) of the car. Actions are discrete,
a throttle of $\{-1,0,1\}$. The agent starts at the position $-0.5$
and a velocity of $0$, and the
goal is at the position $0.6$. The rewards are $-1$ for every time
step. An episode ends when the goal is reached, or when 2000
steps % \footnote{Note the significantly shorter lifetime of an episode here, as compared
  % to results in Degris et al.~\cite{degris2012}; since the shaped rewards are more
  % informative, they can get by with very rarely reaching the goal.}
 have elapsed. The state space is approximated with the standard
tile-coding technique~\cite{sutton-barto98}, using ten tilings of
$10\times 10$, with a parameter vector learnt for each action.

% \begin{figure}[h]
%   \centering
%   \includegraphics[scale=0.4]{mc_mine.png}
%   \caption{\small The mountain car problem. The mountain height $h$ is given
%     by  $h=\sin (3x)$. % Figure from Taylor's dissertation~\cite{thesis-taylor}. 
%     % TODO: ADD CAR, FLAG
%   }
%   \label{fig:mc}
% \end{figure}

In this domain we define three intuitive shaping potentials:

\begin{description}
\item[Position] Encourage progress to the right (in the direction of
  the goal). This potential is flawed by design, since in order to
  get to the goal, one needs to first move away from it.
  \begin{equation}
\label{eq:4}
\Phi_1({\bf x})=% c_r\times 
\bar{x}
\end{equation}
\item[Height] Encourage higher positions (potential
  energy):
  \begin{equation}
    \label{eq:2}
    \Phi_2({\bf x})=% c_h\times 
    \bar{h}
  \end{equation}
\item[Speed] Encourage higher speeds (kinetic energy):
  \begin{equation}
    \label{eq:3}
    \Phi_3({\bf x})=% c_s\times 
    |\bar{\dot{x}}|^2
  \end{equation}
\end{description}

Here ${\bf x} = \langle x,\dot{x}\rangle$ is the state (position and
velocity), and $\bar{a}$ denotes the normalization of $a$ onto $[0,1]$.

We used $\gamma=0.99$. The learning parameters were tuned w.r.t. the
base learner and shared among all demons: $\lambda=0.4,\beta=0.0001,\alpha=0.1$,
where $\lambda$ is the trace decay parameter, $\beta$ the step size for the
second set of weights $w$ in Greedy-GQ, and $\alpha$ the step size for
the main parameter vector $\theta$. We
ran 1000 independent runs of 100 episodes each, with evaluation occuring every 5 episodes ($z=5$).

% We iteratively build up the scenarios to showcase the behavior of our
% framework. We first present the results of the three shapings on their
% own, then an ensemble consisting of the first two.

\begin{figure}[h!]
\centering
   \includegraphics[scale=0.5]{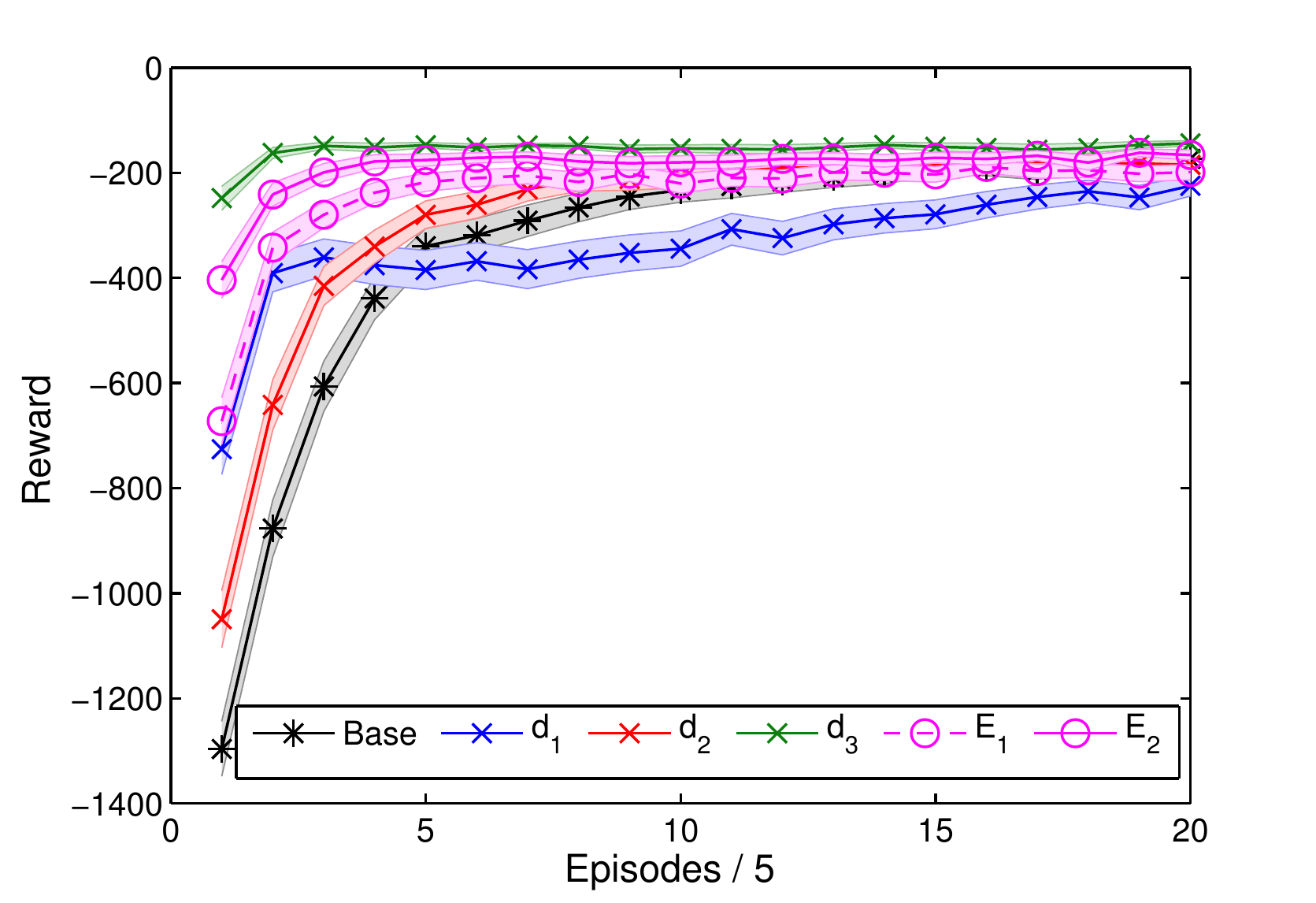}
   \caption{\small Learning curves of the single shapings and their ensembles in
     mountain car. $E_1$, the ensemble of two comparable shapings,
     outperforms both of them, whereas $E_2$, the ensemble of all
     three shapings, matches ($p > 0.05$) the performance of the (more effective)
     third shaping $d_3$.}
\label{fig:learning-curves}
\end{figure}

\subsubsection{Choice of Heuristic}
\label{sec:tuned-ensemble}

% TODO: REWORD SO IT'S VERY CLEAR THAT UNREALISTIC SCENARIO.
In this
experiment\footnote{This experiment first appeared in the early
  version of this work~\cite{harutyunyan2014ecai}.} we address the question of the choice between
heuristics. We thus consider ensembles composed of the demons shaped
with the three shaping potential functions $\Phi_1$,$\Phi_2$ and
$\Phi_3$, and scaled with factors $c_1,c_2,c_3$ that have been tuned beforehand. We
associate the learner $d_i$ with $d^{\Phi_i}_{c_i}$.

When evaluating the shapings individually, we witness $d_3$ to
perform best amongst the three. %  In such a
% case, one would likely prefer to just use that shaping on its own, but
% we remind the reader that this information is not available a
% priori.
To examine the quality of our ensembles w.r.t. the quality of
its components, we consider two scenarios: $E_1=\langle
d_1,d_2\rangle$ of two demons and $E_2=\langle
d_1,d_2,d_3\rangle$ of three
demons. This corresponds to having ensemble consisting of two
comparable shapings, and an ensemble with one clearly most efficient
shaping. % DOMINATING VS COMPLEMENTARY 
Thus, ideally, we would like
$E_1$ to outperform both $d_1$ and $d_2$ and $E_2$ to at least match
the performance of $d_3$.

Fig.~\ref{fig:learning-curves} presents the learning performance of
the base agent, the demons $d_1,d_2,d_3$ shaped with single potentials, and
the two ensembles $E_1$ and $E_2$, mentioned above. We witness the individual shapings alone to
aid the learning significantly. $E_1$ follows $d_1$ at first,
when its performance is better, but switches to $d_2$, when
the performance of $d_1$ levels out. This
is because $d_1$ (as is appropriate with its position shaping)
persists on going right in the beginning of an episode, and this
strategy, while effective at first, results in a plateau of a higher number of steps. The ensemble policy
is able to avoid this by incorporating information from $d_2$. 

$E_2$, the ensemble of all three shapings, begins better than
both $d_1$ and $d_2$, but slightly worse than $d_3$, the most
effective shaping. It, however, quickly catches up to $d_3$, with the
overall performance of $E_2$ and $d_3$ being statistically indistinguishable.

Thus, the performance of the ensembles
meets our desiderata: when there is clearly a best component, an
ensemble statistically matches it, otherwise it outperforms all of its components.

% and Tables~\ref{tab:sc1} and
% \ref{tab:sc2} give the learning performance of the base learner, the
% individual shaped demons $d_1,d_2,d_3$ and the two ensembles $E_1$ and
% $E_2$. Individual shapings
% alone aid learning speed significantly. The combination method meets
% our desiderata: it either statistically matches
% or is better than the best shaping at any stage, overall outperforming
% all single shapings.

\subsubsection{Choice of Scale}
\label{sec:shap-mult-scal}

The previous set of experiments assumed access to the best scaling
factors $c_1,c_2,c_3$. In practice obtaining these requires tuning
each shaping prior to the use of the ensemble, a scenario we aim to
avoid. In this section we demonstrate that ensembles on a
range of scales perform at least as well, as those with
cherry-picked components.

\begin{figure}[h!]
  \centering
  \includegraphics[scale=0.47]{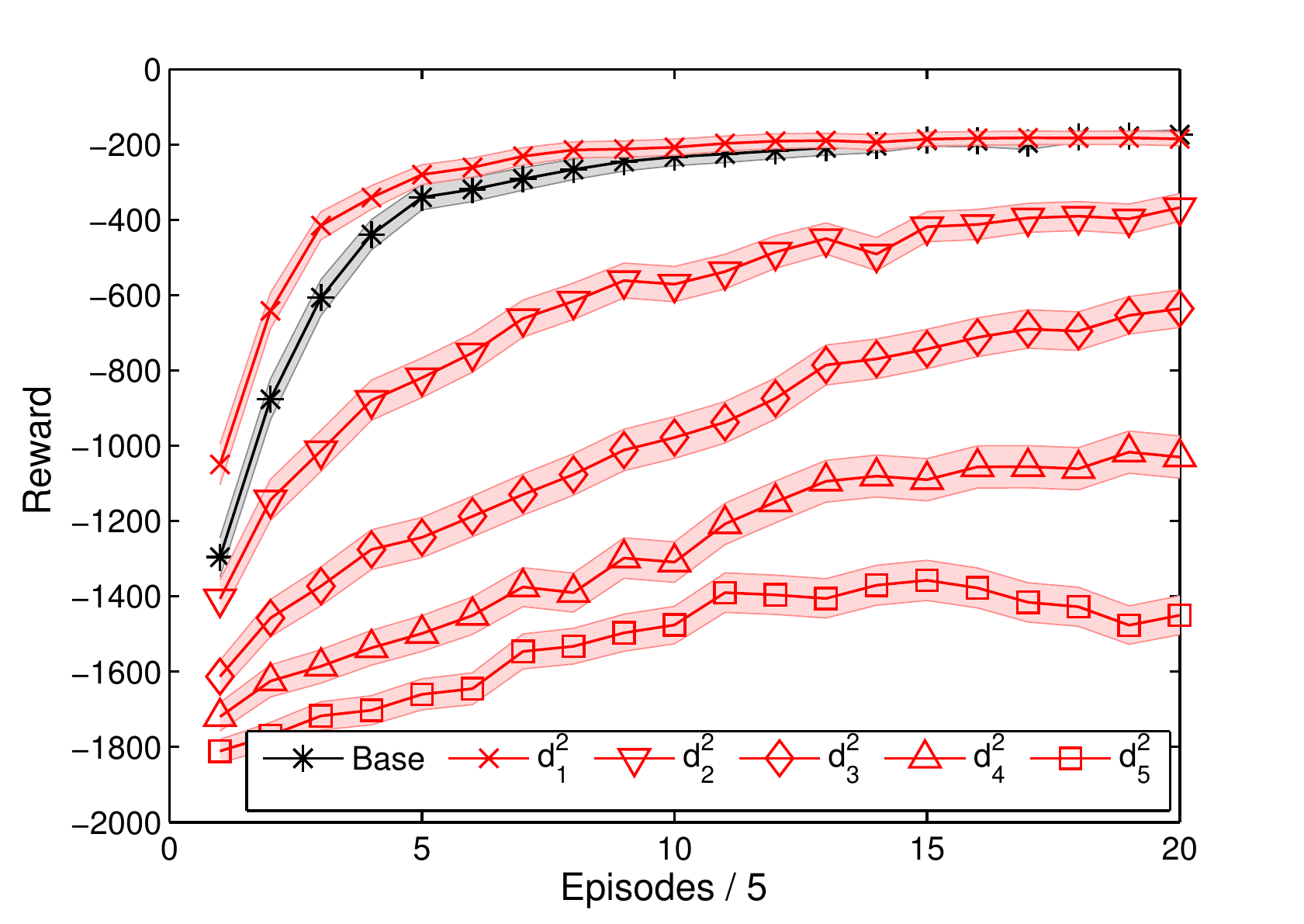}
  \caption{\small The range of performance of a single shaping
    w.r.t. different scales in mountain car. Each curve corresponds to the performance
    of a demon shaped with $\Phi_2$, with a scaling
    factor from the range $C_1$.}
  \label{fig:scale-effect}
\end{figure}

Namely, we consider two
scaling ranges $C_1=\langle 20,40,60,80,100\rangle$ and
$C_2=\langle 1,10,10^2,10^3,10^4\rangle$, with the first being a reasonably close range
to the optimal scales from the previous section, and the second being a general sweep, with
no intuition or knowledge of the optimal scale. Before we proceed
further, we illustrate the effect a scaling factor can have on the
performance of a single shaping. Fig.~\ref{fig:scale-effect} gives a
comparison of the performance of the shaping potential $\Phi_2$ over
the (reasonable) scaling range $C_1$. Even small differences in
scale have dramatic effect on the shaping's performance.

% The scalability of Horde allows to learn on a {\em range} of scaling factors for each shaping,
% and, as before, deduce the ensemble policy by voting.

Now let $E_{C_1}$ and $E_{C_2}$ be the ensembles w.r.t. all three
shapings on $C_1$ and $C_2$, resp., each totaling in 16 demons
(including the base learner). We compare $E_{C_1}$ and $E_{C_2}$ with
$E_2$ (the ensemble w.r.t. the three
shapings with tuned scaling factors, from the first experiment).  We
illustrate the range of performances of shapings for each scale range,
by additionally plotting the {\em average} of the runs of each shaping
across each scale. I.e. for the range $C_j$, and shaping $\Phi_i$, at
each episode, this is the average of the rewards obtained by the
demons $d^i_1$, $d^i_2$,$\ldots$,$d^i_{|C_j|}$ in that episode.

% To illustrate the range of
% performances of shapings across the range of scales, we also include the plots
% of the {\em average} for each shaping across $C_1$ and $C_2$ (i.e. an
% average of the runs of 5 demons each).

\begin{figure}[h!]
\centering
\includegraphics[scale=0.47]{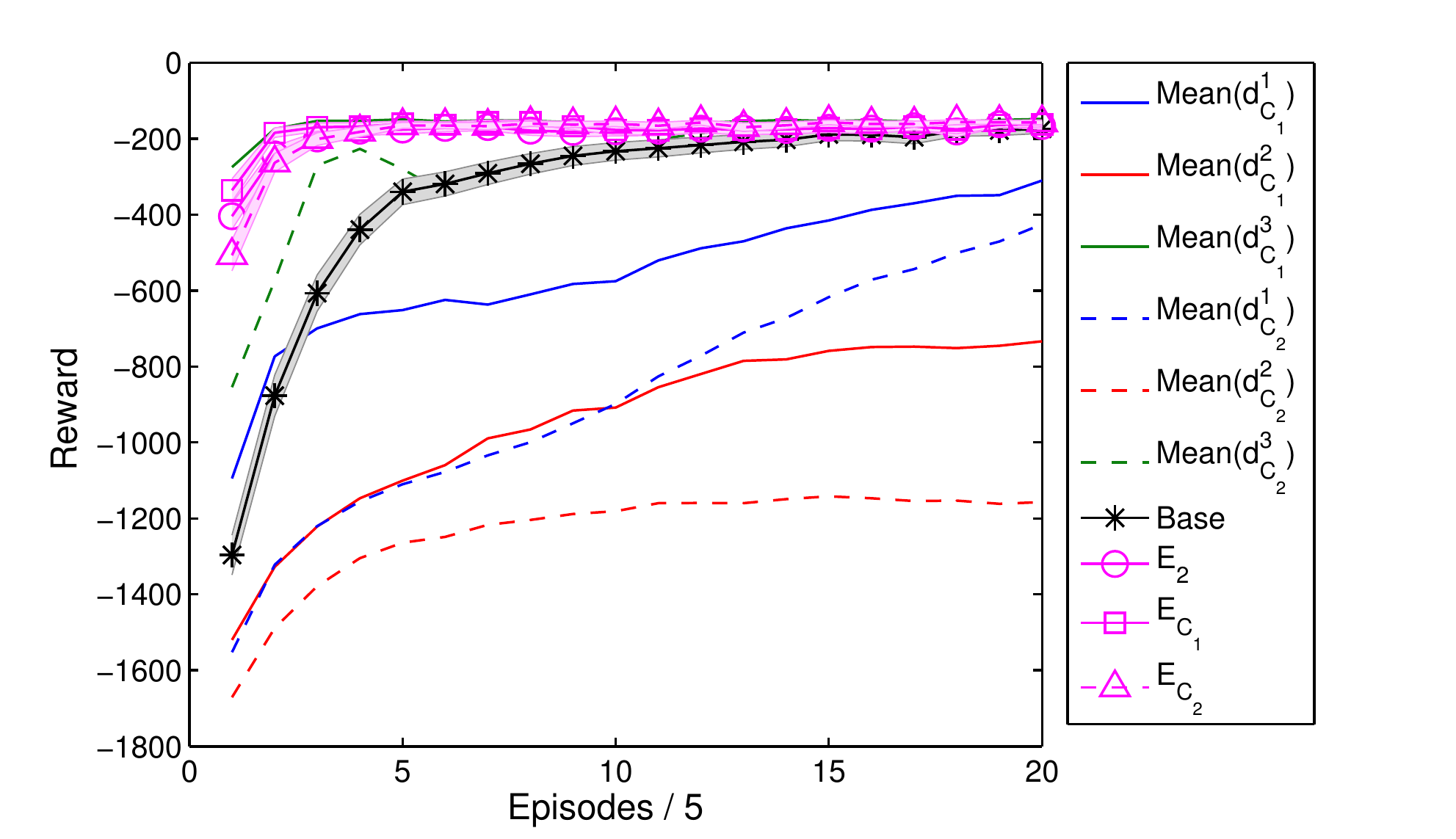}
\caption{\small Learning curves of the ensembles over the scale
     ranges $C_1$ and $C_2$ in mountain car. The solid and dashed lines (for each of
     the three shapings) are the {\em
       mean} performance of the demons w.r.t. $C_1$ and $C_2$,
     respectively, and are plotted as a reference for the performance
     of the ensemble components. Note that there is no single demon
     with this performance. The performances of ensembles $E_{C_1}$ and $E_{C_2}$ are not
     significantly different from that of $E_2$: the ensemble
     w.r.t. tuned components.}
\label{fig:learning-curves-ens}
\end{figure}

Fig.~\ref{fig:learning-curves-ens} % and Table~\ref{tab:mc} 
presents the results. $E_{C_1}$ and $E_{C_2}$ are both statistically
the same ($p>0.05$) as the tuned ensemble $E_2$, despite their components having a much wider range of performance. 

% \begin{table}
% \begin{center}
% {\caption{Mountain car. Average of 1000 independent runs of 100
%     episodes each. Performance is given as the mean reward over a run
%     of the 100 episodes, and indicared with a standard deviation. The
%     mean performance of a shaping is the mean of the means, and the
%     standard deviation is w.r.t. the outer mean.}\label{tab:sc1}}
% \begin{tabular}{|  l c | c  | c |} \hline
%   \multicolumn{2}{ | c| }{Variant} & \multicolumn{2}{c |}{Scaling range}\\
%   & & $C_1$ & $C_2$ \\ \hline
%   \multicolumn{2}{|c|}{Base} & \multicolumn{2}{|c|}{-343.7$\pm$283.0} \\ \hline
%   \multirow{3}{*}{Position $\Phi_1$} & {\em mean} & -550.4$\pm$231.15 & -871.3$\pm$535.3\\
%   & {\em best} & -333.3$\pm$223.0 & -338.0$\pm$257.4 \\
%   & {\em worst} & -856.3$\pm$279.9 & -1432$\pm$453.4 \\ \hline
% \multirow{3}{*}{Height $\Phi_2$} & {\em mean} & -944.1$\pm$493.9 &
% -1234.4$\pm$848.4 \\
%   &{\em best} & -286.3$\pm$211.2 & -318.3$\pm$276.7 \\
%   & {\em worst} & -1521.4$\pm$144.9 & -1997.5$\pm$1.1 \\ \hline
% \multirow{3}{*}{Speed $\Phi_3$} & {\em mean} & -160.8$\pm$6.5 & -265.6$\pm$82.6 \\
%   & {\em best} & -156.6$\pm$22.0 & -160.6$\pm$27.4 \\
%   & {\em worst} & -172.2$\pm$42.3 & -365.2$\pm$309.7 \\ \hline
% \multicolumn{2}{|c|}{Ensemble $E_C$} & {\bf -177.9$\pm$37.9} & {\bf -187.18$\pm$79.0} \\ \hline
% \multicolumn{2}{|c|}{Tuned ensemble $E_2$} &
% \multicolumn{2}{|c|}{\bf -190.3$\pm$53.0} \\ \hline
% \end{tabular}
% \end{center}
% \label{tab:mc}
% \end{table}

\subsection{Cart-Pole}
\label{sec:cartpole}

We now validate our framework on the problem of cart-pole~\cite{michie:boxes}. The task is
to balance a pole on top of a moving cart for as long as possible. The (continuous)
state $\bf s$ contains the angle $\xi$ and angular velocity $\dot{\xi}$ of the pole, and the
position $x$ and velocity $\dot{x}$ of the cart. There are two
actions: a small positive and a small negative force applied to the
cart. A pole falls if $|\xi|>\frac{\pi}{4}$, which terminates the
episode. The track is bounded within $[-4,4]$, but the sides are
``soft''; the cart does not crash upon hitting them. The
reward function penalizes a pole drop, and is 0 elsewhere. An episode
terminates successfully, if the pole was balanced for 1000 steps. The
state space is approximated with tile coding, using ten tilings of
$10\times 10$ over all 4 dimensions, with a parameter vector learnt
for each action. % Like in mountain car, the behavior policy is a
% uniform distribution over all actions at each time step.

We define two potential functions, corresponding to the angle and
angular speed of the pole.

\begin{description}
\item[Angle] Discourage angles far from the equilibrium:
  \begin{equation}
\label{eq:4}
\Phi_1({\bf s})=-\bar{|\xi|}^2
\end{equation}
\item[Angular speed] Discourage high speeds (which are
  likelier to result in dropping the pole):
  \begin{equation}
    \label{eq:2}
    \Phi_2({\bf s})=-|\bar{\dot{\xi}}|^2
  \end{equation}
\end{description}

We used $\gamma=0.99$. The learning parameters were tuned w.r.t. the
base learner and set to $\lambda=0.7$, $\alpha=0.1$ and
$\beta=0.001$. These settings were shared among all demons.
We ran 100 independent runs of a 1000 episode each,
with evaluation occuring every 50 episodes ($z=50$).

\subsubsection{Choice of Heuristic and Scale}
\label{sec:choice-heur-scale}

In this experiment we evaluate the problems of the choice of the
heuristic and its scale jointly. We consider a general scaling range $C=\langle
1,10,10^2,10^3,10^4\rangle$, and three ensembles: $E^1_C$ resp. $E^2_C$ only comprised
of the demons shaped w.r.t. $\Phi_1$ resp. $\Phi_2$ across $C$ (5 demons each),
and $E_C$ containing all 11 demons (including the base
learner). As before, we illustrate the range of performances of
shapings across the range of scales by, for each shaping, plotting the {\em average}
performance of the demons w.r.t. that shaping across the entire
scale range. I.e. for the shaping $\Phi_i$, at each episode, this is
the average of the rewards obtained by the demons $d^i_1$, $d^i_2$,$\ldots$,$d^i_{|C|}$ in that episode.

% To illustrate the range of performances of shapings across the range of scales, we also include the plots
% of the {\em average} for each shaping across $C$. 

\begin{figure}[h!]
  \centering
  \includegraphics[scale=0.4]{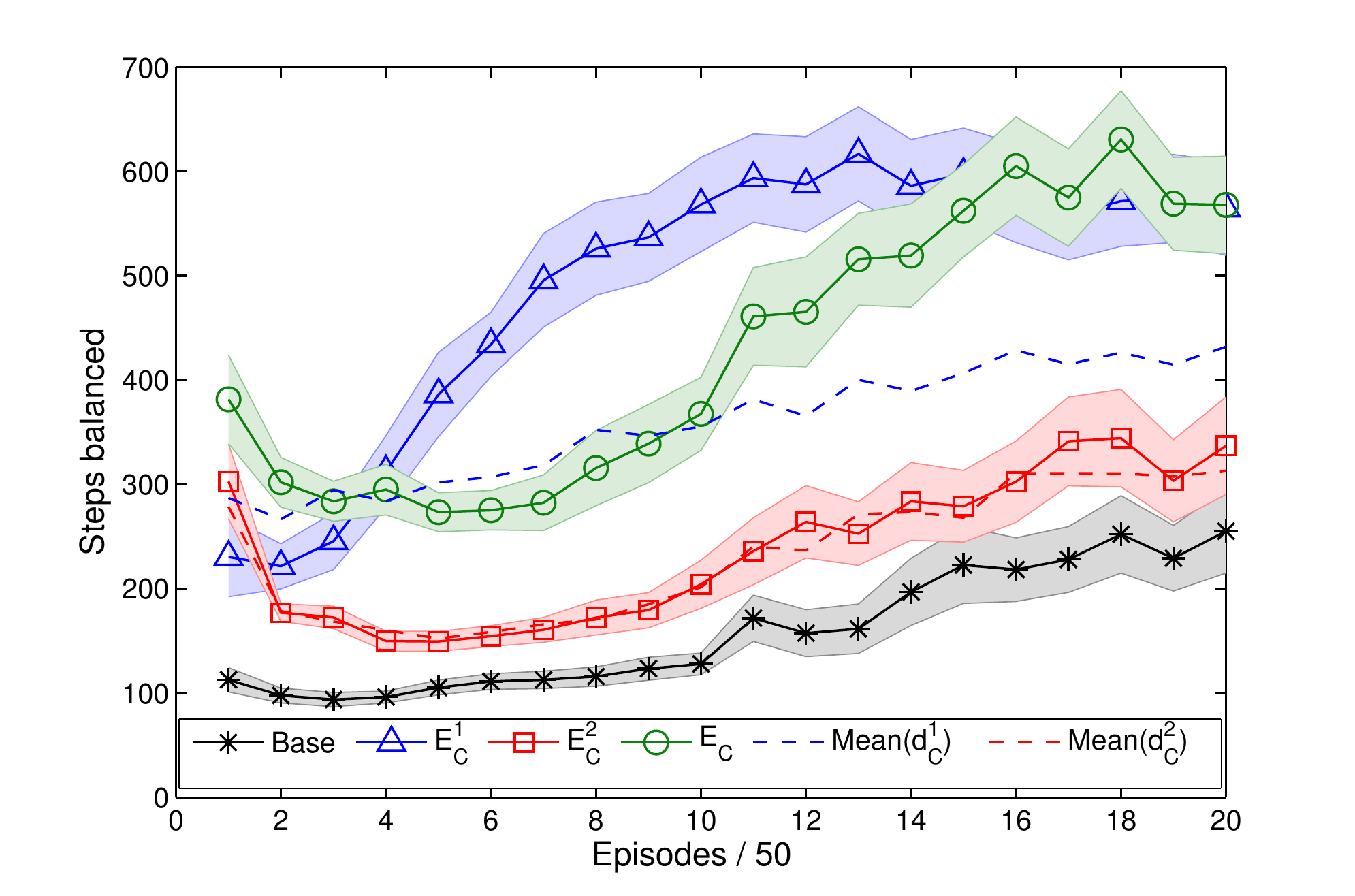}
  \caption{\small Learning curves for the ensembles $E^1_C$, $E^2_C$
    and $E_C$ in cart-pole. The dashed lines (for each of
     the two shapings) denote the {\em mean} performance of the demons
     w.r.t. $C$, and plotted as a reference for the performance of the ensemble components. Note that there is no single demon
     with this performance. The performances of the global ensemble
     $E_C$ follows the (more effective) first shaping, in the end matching the
     performance of the corresponding ensemble $E^1_C$.}
  \label{fig:cp}
\end{figure}

% \begin{table}
% \begin{center}
% {\caption{Cart-pole. Average of 100 independent runs of 1000
%     episodes each. }\label{tab:sc1}}
% \begin{tabular}{|  l c | c  |} \hline
%   \multicolumn{2}{ | c| }{Variant} & Scaling range\\
%   & & $C_1$ \\ \hline
%   \multicolumn{2}{|c|}{Base} & 159.5$\pm$57.3 \\ \hline
%   \multirow{4}{*}{Angle $\Phi_1$} & mean & 358.7$\pm$176.2 \\
%   & best & 563.7$\pm$31.5\\
%   & worst & 162.8$\pm$60.2 \\ 
%   & {\bf ensemble} $E^1_C$  & 489.4$\pm$134.4 \\ \hline
% \multirow{4}{*}{Angular speed $\Phi_2$} & mean & 233.1$\pm$94.7 \\
%   & best & 387.4$\pm$147.2 \\
%   & worst & 166.0$\pm$32.1 \\
%   & {\bf ensemble} $E^2_C$ & 238.4$\pm$70.5\\ \hline
% \multicolumn{2}{|c|}{Ensemble $E_C$} & 429.4$\pm$130.0\\ \hline
% \end{tabular}
% \end{center}
% \label{tab:cp}
% \end{table}

Fig.~\ref{fig:cp} % and Table~\ref{tab:cp} 
shows the results. All
ensembles (and ensemble averages) improve over the base learner. The performance of $E^2_C$, the ensemble over the second shaping, matches that of
the average from that ensemble, since all of its components perform
similarly. On the other hand, $E^1_C$, the ensemble over the first
shaping, does much better than the corresponding average. The global ensemble $E_C$
over all of the demons starts out better than both $E^1_C$ and
$E^2_C$, then levels at the average performance of the (better) first
shaping, and finally matches the performance of $E^1_C$. The global
ensemble $E_C$ thus correctly identifies both {\em which shaping} to follow: its
performance always follows (or is better than) that of the more
efficient first shaping (either on average, or the ensemble $E^1_C$),
and on {\em what scales}: the final performance of $E_C$ matches that of
$E^1_C$, significantly improving over the average across the scale range.

\section{Conclusions}
\label{sec:concl-future-work}

In this work we described a novel off-policy PBRS ensemble architecture that is able
to improve learning speed in a latent setting, without requiring the extra sample
complexity introduced by the steps of tuning the heuristic
and its scale, typical to PBRS. We avoid these steps by learning an
ensemble of policies w.r.t. many heuristics and scaling factors
simultaneously. Our ensemble possesses
general convergence guarantees, while staying efficient, as it leverages
the recent Horde architecture to learn a single task well. Our
experiments validate the use of PBRS in the latent setting, and
demonstrate the efficacy of the proposed ensemble. Namely, we show
that the ensemble policy over both broad and narrow ranges of scales performs at
least as well as the one over a set of optimally pre-tuned components,
which in turn performs at least as well as its best
component-heuristic. 

% In this paper our focus has been a latent learning scenario, in which
% the behavior policy is fixed. While this is an important case,
% relaxing the behavioral constraint would further expand the
% applicability of the architecture. Extending the guarantees provided
% by Horde to more flexible behaviors is a topic of ongoing research in
% the GTD community.

% We gave the first policy ensemble that is both sound and
% capable of learning in real time, by exploiting the power of Horde
% architecture to learn a single policy well. 
% The value functions in our ensemble learned on shaped
% rewards, and we used a voting method to combine them.  

% We validated the approach on two domain, and witnessed
% performance benefits in both the case with the ensemble consisting of
% tuned component shapings, and one over general scaling ranges.
% considering two scenarios: with and without a clearly best shaping
% signal. In the former scenario, the combination outperformed single
% shapings, and in the latter was able to match the performance of that
% best shaping.
% IT OUTPERFORMED AND KICKED ASS BLAH.
% In general, we expect to see larger benefits on larger problems; a
% more extensive suite of experiments is subject to future work.

% The primary limitation of Horde is the requirement to keep the
% behavior policy fixed (or change it slowly). While this is an
% important case, relaxing this constraint would further expand the
% effectiveness of the architecture. This is a topic of ongoing research in the GTD community.

\subsection*{Future Directions}
\label{sec:future-work}

In this work we have assumed a shared set of parameters between the demons, an immediate
extension would be to maintain demons that learn w.r.t. different
parameters. This is similar to the
approach of Marivate  and Littman~\cite{marivate2013}, who learn to
solve many variants of a problem for the best parameter settings in a
generalized MDP. In our case the MDP (dynamics) will remain shared, but the
individual parameters of the demons will vary. 

It would be worthwhile to evaluate the framework w.r.t. different ensemble
techniques that induce the target ensemble policy. This would be
especially useful in domains where only select scaling factors of
select heuristics offer improvement: taking a global majority vote over such
an ensemble will likely not be as effective, as trying to determine
which subset of demons to consider. One could, e.g., use
confidence measures~\cite{brys2014combining} to identify these demons.

% The effectiveness of off-policy reward shaping we witness in this
% paper is of both theoretical and practical interest.

Instead of shaping demons with static potential functions, one could
consider maintaining a layer of demons that each {\em learn} some
potential function~\cite{marthi2007,grzes2010}, which are, in turn,
fed into the layer of shaped demons who contribute to the ensemble
policy. One needs to be realistic about attainability of learning this
in time, since as argued by Ng et al.~\cite{ng99}, the best potential
function correlates with the optimal value function $V^*$, learning
which would solve the base problem itself and render the potentials
pointless.

% In this work, we considered an ad-hoc voting approach to combining
% shapings. One of the possible future directions would be to {\em learn}
% optimal combination ways via predicting some shared fitness value
% w.r.t. the policies induced by the learnt value functions. The
% challenge with this is that the meta-learning has to happen at a much
% faster pace for it to be useful in speeding up the main learning
% process. In the case of shapings, this is doubly the case, since they
% all eventually converge to the same (optimal) policy. The size of this window
% of opportunity is related to the size of the problem.

% One could go further and attempt to {\em
%   learn} the best potential functions~\cite{marthi2007,grzes2010}. As
% before, one needs to be realistic about attainability of learning this in
% time, since as argued by Ng et al.~\cite{ng99}, the best potential
% function correlates with the optimal value function $V^*$, learning
% which would solve the base problem itself and render the potentials
% pointless.

% \section{Acknowledgments}
% Anna Harutyunyan is supported by the IWT-SBO
% project MIRAD (grant nr. 120057). Tim Brys is funded by a Ph.D grant of the Research Foundation-Flanders (FWO).

%\newpage
\bibliographystyle{abbrv}
\bibliography{all} 

\end{document}